%% file: example_paper.tex
\documentclass{article}

\usepackage{microtype}
\usepackage{graphicx}
\usepackage{booktabs} 
\usepackage{caption}

\usepackage{float} 
\usepackage{placeins}  
\usepackage{dblfloatfix}  

\usepackage[accepted]{icml2025}

\usepackage{hyperref}

\usepackage{multirow}
\usepackage{colortbl} 
\usepackage{adjustbox}
\usepackage{listings}
\usepackage{amsmath}
\usepackage{amssymb}
\usepackage{mathtools}
\usepackage{amsthm}
\usepackage{subcaption}
\usepackage{enumitem}

\usepackage[capitalize,noabbrev]{cleveref}

\theoremstyle{plain}

\theoremstyle{definition}

\theoremstyle{remark}

\usepackage[textsize=tiny]{todonotes}

\icmltitlerunning{DLaVA}

\begin{document}

\twocolumn[
\icmltitle{DLaVA: Document Language and Vision Assistant for Answer Localization with Enhanced Interpretability and Trustworthiness 
}



\icmlsetsymbol{equal}{*}

\begin{icmlauthorlist}
\icmlauthor{Ahmad Mohammadshirazi}{yyy}
\icmlauthor{Pinaki Prasad Guha Neogi}{yyy}
\icmlauthor{Ser-Nam Lim}{comp}
\icmlauthor{Rajiv Ramnath}{yyy}
\end{icmlauthorlist}

\icmlaffiliation{yyy}{Department of Computer Science and Engineering, Ohio State University, Ohio, US}
\icmlaffiliation{comp}{Department of Computer Science, University of Central Florida, Florida, US}

\icmlcorrespondingauthor{Ahmad Mohammadshirazi}{mohammadshirazi.2@osu.edu}

\icmlkeywords{}

    

\vskip 0.3in
]



\printAffiliationsAndNotice{\icmlEqualContribution} 


\input{sec/0_abstract}     
\input{sec/1_intro}

\input{sec/2_relatedwork}

\input{sec/3_method}

\input{sec/4_experiment}
\input{sec/5_result}

\nocite{langley00}

\bibliography{example_paper}
\bibliographystyle{icml2025}

\newpage
\appendix

\input{sec/append}

\end{document}

%% file: sec/0_abstract.tex
\begin{abstract}

Document Visual Question Answering (VQA) demands robust integration of text detection, recognition, and spatial reasoning to interpret complex document layouts. In this work, we introduce DLaVA, a novel, training-free pipeline that leverages Multimodal Large Language Models (MLLMs) for zero-shot answer localization in order to improve trustworthiness, interpretability, and explainability. By leveraging an innovative OCR-free approach that organizes text regions with unique bounding box IDs, the proposed method preserves spatial contexts without relying on iterative OCR or chain-of-thought reasoning, thus substantially reducing the computational complexity. We further enhance the evaluation protocol by integrating Intersection over Union (IoU) metrics alongside Average Normalized Levenshtein Similarity (ANLS), thereby ensuring that not only textual accuracy is considered, but spatial accuracy is taken into account, ultimately reducing the risks of AI hallucinations and improving trustworthiness. 
Experiments on benchmark datasets demonstrate competitive performance compared to state-of-the-art techniques, with significantly lower computational complexity and enhanced accuracies and reliability for high-stakes applications. The code and datasets utilized in this study for DLaVA are accessible at: https://github.com/ahmad-shirazi/AnnotMLLM.

\end{abstract}

%% file: sec/1_intro.tex
\vspace{-5mm}
\section{Introduction}

\label{sec:intro}

\begin{figure}[h!]
    \centering
    \begin{subfigure}{0.25\textwidth}
        \centering
        \includegraphics[width=\linewidth, trim={0cm 4.5cm 0cm 0cm}, clip]{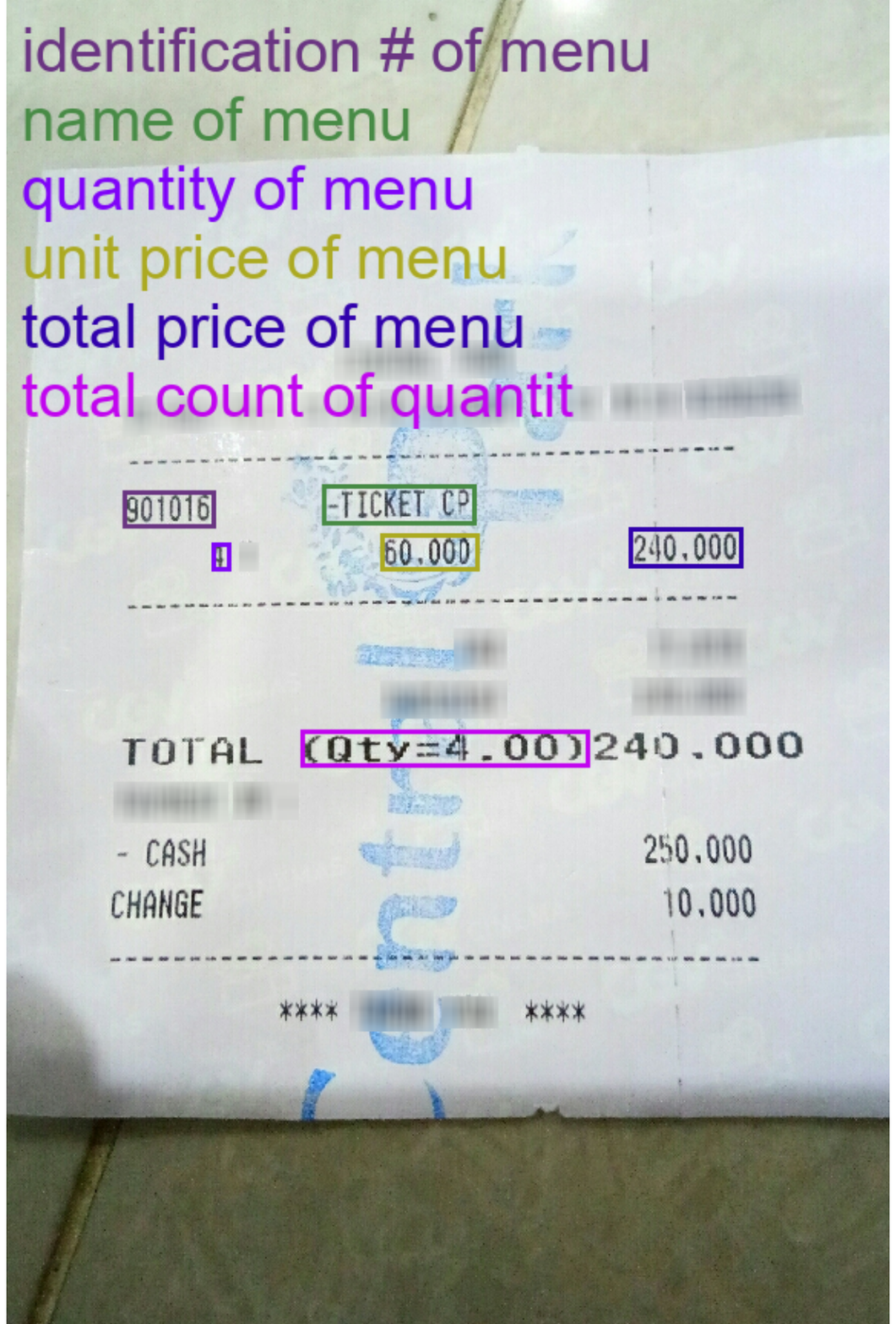}
    \end{subfigure}%
    \hspace{-\linewidth}
    \begin{subfigure}{0.25\textwidth}
        \centering
        \includegraphics[width=\linewidth, trim={0cm 4.5cm 0cm 0cm}, clip]{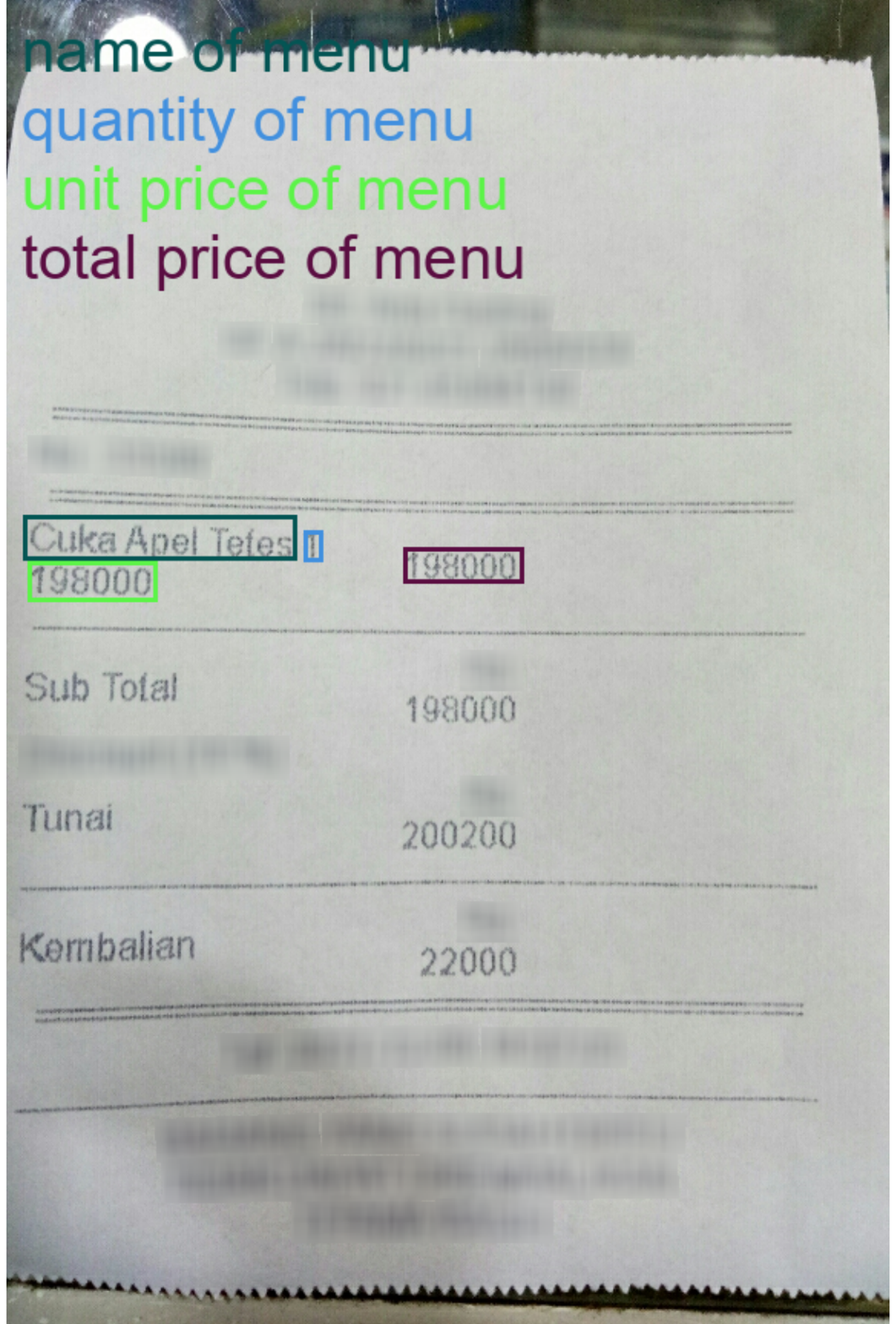}
    \end{subfigure}
    
    \caption{Examples of visual information extraction on images from the CORD dataset \cite{park2019cord}: questions are displayed at the top in colored fonts, with the corresponding answers highlighted by matching colored boundary boxes.} 

    \label{fig:sample}
\end{figure}


    

\begin{figure*}[h]
    \centering
    \includegraphics[width=1\textwidth]{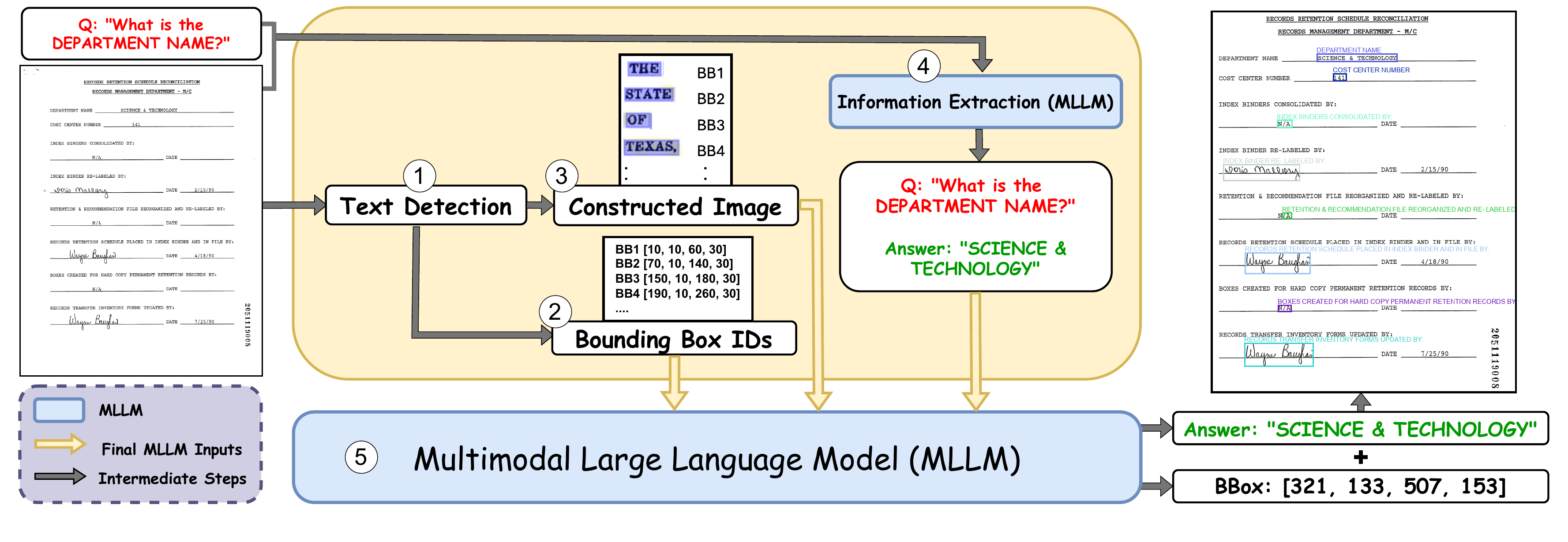}
    \caption{DLaVA Model Architecture. This diagram illustrates our final single-pipeline design. In the text detection step, detected text regions generate two outputs: a series of cropped images that are reorganized into a “constructed image” with unique bounding box identifiers (e.g., BB1, BB2, BB3, etc.) and their corresponding bounding box coordinates (e.g., BB1 [10, 10, 60, 30], BB2 [70, 10, 140, 30], etc.). The approach then leverages a two-stage MLLM pipeline. In Stage 1, the original image and the user’s question are provided to the MLLM to derive an initial textual answer. In Stage 2, the constructed image—comprising all cropped images with their BB IDs—along with the recorded bounding box coordinates and the initial QA pair are fed back into the MLLM to refine spatial localization. This integrated design eliminates the need for iterative OCR and reduces computational overhead, culminating in a final annotation module that delivers the final answer along with precise bounding box annotations. Numbered circles denote sequential processing steps (see Section~\ref{sec:method} for more details).}
    
    
    \label{fig:anot}
\end{figure*}

Document Visual Question Answering (VQA) stands at the intersection of computer vision and natural language processing, aiming to answer questions based on the content of a document image. This task is inherently challenging due to the need for a model to not only accurately recognize and interpret textual information within complex visual layouts but also to reason about the spatial relationships and semantics of the content. Effective solutions require a harmonious integration of text detection, recognition, and contextual understanding to bridge the gap between visual data and linguistic queries \cite{ishmam2024image}. Figure \ref{fig:sample} presents some examples of visual information extraction, showcasing document annotations from the CORD dataset (see Appendix \ref{appendix:Examples-Ground-Truth-Answer-Annotations} and \ref{appendix:Examples-Predicted-Answer-Annotations} for more details).

Existing approaches, such as LayoutLMv3 \cite{huang2022layoutlmv3}, LayoutLLM \cite{luo2024layoutllm}, LayTextLLM \cite{lu2024bounding}, and DocLayLLM \cite{liao2024doclayllm}, have made significant progress in visual question answering and layout analysis. However, these methods come with several limitations. They often rely on chain-of-thought (CoT) reasoning or iterative OCR processes for spatial grounding, which incur high computational costs and require extensive fine-tuning. Furthermore, these methods are evaluated on metrics like Average Normalized Levenshtein Similarity (ANLS) \cite{yujian2007normalized} that focus primarily on textual accuracy while overlooking the spatial correctness of the predicted answers. As a result, these approaches typically lack precise answer localization, thereby limiting interpretability and explainability—challenges that are particularly critical in high-stakes applications such as legal, medical, and financial document analysis \cite{huang2024trustllm}.

Motivated by these challenges, we propose a novel, zero-shot (training-free) OCR-free pipeline that harnesses the inherent visual understanding of Multimodal Large Language Models (MLLMs) to directly extract and localize answers from document images. Unlike conventional OCR-dependent methods—which often suffer from cascading errors and high computational complexity—our approach bypasses the need for iterative OCR by constructing a \emph{single} image that comprises detected text regions with unique bounding box identifiers, thereby preserving essential spatial relationships while significantly reducing computational overhead. 

Our approach begins with a text detection module that produces two key outputs: (i) a series of cropped images from detected text regions that are reorganized into a “constructed image” with unique bounding box identifiers (e.g., BB1, BB2, BB3, etc.), and (ii) the corresponding bounding box coordinates (e.g., BB1 [10, 10, 60, 30], BB2 [70, 10, 140, 30], etc.). In the first stage of our two-stage pipeline, the original image and the user’s question are provided to the MLLM to generate an initial textual answer. In the second stage, the constructed image—comprising all cropped images with their BB IDs—along with the recorded bounding box coordinates and the initial QA pair, are fed back into the MLLM to refine spatial localization, culminating in a final annotation module that outputs the final answer with precise bounding box annotations on the input image. 

This design is driven by several key motivations: firstly, by consolidating text information into a single constructed image rather than sending all recognized text as prompts, typical in OCR-dependent methods, we reduce the token count, which is crucial for avoiding context window overflow (e.g., the 128k token limit for Pixtral) and ensuring that the MLLM can process the input effectively; secondly, the constructed image approach bypasses the iterative OCR processing required for each cropped image, thereby streamlining the pipeline and reducing computational overhead; and finally, instead of processing multiple separate cropped images—which may exceed the MLLM’s input limitations—we combine them into a single constructed image, making the model more efficient and suitable for spatial reasoning. We demonstrate the effectiveness of our model by comparing it with an OCR-dependent baseline, and empirical evaluations confirm that our model not only attains state-of-the-art (SoTA) textual accuracy but also achieves robust spatial grounding, establishing its potential as a viable alternative to CoT or OCR-dependent solutions. Building on this foundation, our contributions are as follows:

\begin{enumerate}
    \item \textbf{Zero-shot Answer Localization in MLLMs:} We introduce a novel pipeline that augments MLLMs with the ability to localize answers within document images using a training-free, zero-shot paradigm. This approach significantly reduces computational complexity compared to traditional CoT or fine-tuning methods.
    
    \item \textbf{Innovative Pipeline Design with a Constructed Image:} We propose a streamlined pipeline that integrates an MLLM with a text detection module, eliminating the need for a separate text recognition component. Our key innovation is the concept of a ``constructed image'' (see Figure 2), where text regions are organized with unique bounding box IDs to preserve spatial relationships. This design not only simplifies the processing pipeline but also delivers superior performance on benchmark datasets. 
    

    \item \textbf{Enhanced Interpretability \& Explainability through Comprehensive Evaluation with IoU Metrics:} Our method improves model transparency by providing spatially grounded responses. By incorporating Intersection over Union (IoU) \cite{rezatofighi2019generalized} metrics alongside ANLS, we offer a rigorous evaluation framework that captures both textual and spatial accuracy, thereby enhancing interpretability and reliability. Furthermore, by annotating and pinpointing the precise locations of predicted answers, our approach reduces the risk of AI hallucinations, ensuring that the model’s outputs remain firmly grounded in the visual evidence presented in the document.

\end{enumerate}



\noindent The remainder of this paper is organized as follows: In Section~\ref{sec:related}, we review related work and present a literature survey. Section~\ref{sec:method} details our proposed approach - DLaVA. Section~\ref{sec:Experiments} presents the experimental setup, including dataset descriptions, hyperparameters, baseline models, and ablation study models. In Section~\ref{sec:result}, we discuss the results, highlighting the trustworthiness, interpretability, and explainability of the proposed DLaVA model. Finally, Section~\ref{sec:conclusion} concludes the paper and outlines the future work.

%% file: sec/2_relatedwork.tex
\begin{table*}[htbp]
\centering
\setlength{\tabcolsep}{2pt}
\caption{Comparison of DLaVA with SoTA models on benchmark datasets using ANLS evaluation metric}
{\small
\begin{tabular}{llcccccc}
\toprule
\multirow{2}{*}{\textbf{Model Category}} & \multirow{2}{*}{\textbf{Models}} & \multicolumn{3}{c}{\textbf{Document VQA}} & \multicolumn{3}{c}{\textbf{QA for VIE}} \\
\cmidrule(lr){3-5} \cmidrule(lr){6-8}
& & \textbf{DocVQA} & \textbf{EST-VQA} & \textbf{RICO} & \textbf{FUNSD} & \textbf{CORD} & \textbf{SROIE} \\

\midrule

\multirow{2}{*}{\textbf{Text}} 

& Llama2-7B-Chat \cite{touvron2023llama} & 64.99 & 52.14 & 59.49 & 48.20 & 47.70 &68.97  \\
& Llama3-8B-Instruct \cite{dubey2024llama} & 51.79 & 54.65 & 58.81 & 68.57 & 52.31 &61.24 \\

\midrule
\multirow{1}{*}{\textbf{Text + BBox }} 
& LayTextLLM \textbf{(Llama2-7B)} \cite{lu2024bounding} & 72.83 & - & - & 78.65 & 70.81 & 83.27 \\

\midrule
\multirow{4}{*}{\textbf{Text + BBox + Image}} 
& LayoutLLM-7B \textsubscript{CoT} \textbf{(Llama2-7B)} \cite{luo2024layoutllm} & 74.25 & - & - & 78.65 & 62.21 & 70.97 \\
& LayoutLLM-7B \textsubscript{CoT} \textbf{(Vicuna-1.5-7B)} \cite{luo2024layoutllm} & 74.27 & - & - & 79.98 & 63.10 & 72.12 \\

& DocLayLLM \textbf{(Llama2-7B)} \cite{liao2024doclayllm} & 72.83 & - & - & 78.65 & 70.81 & 83.27 \\
& DocLayLLM \textbf{(Llama3-7B)} \cite{liao2024doclayllm} & 78.40 & - & - & 84.12 & 71.34 & 84.36 \\

\midrule
\midrule

\multirow{6}{*}{\textbf{Image}} 
& Phi4-14B \cite{abdin2024phi} & 79.84 & 60.22 & 68.49 & 77.64 & 77.03 & 80.12 \\
& Llama3.2-11B \cite{dubey2024llama} & 78.4 & 48.14 & 53.47 & 65.02 & 42.96 & 61.42 \\
& Pixtral-12B \cite{agrawal2024pixtral} & 80.71 & 61.67 & 70.31 & 78.26 & 79.08 & 82.24 \\
& LLaVA-NeXT-13B \cite{liu2023improved} & 51.01 & 13.77 & 25.12 & 19.71 & 33.5 & 13.41 \\
& LLaVA-OneVision-7B \cite{li2024llava} & 47.59 & 22.39 & 19.54 & 22.82 & 32.43 & 12.10 \\
& Qwen2.5-VL-7B \cite{bai2025qwen2} & 68.54 & 61.41 & 56.42 & 58.44 & 39.01 & 56.37 \\
& InternVL2-8B \cite{chen2024internvl} & 71.26 & 59.74 & 44.81 & 57.58 & 55.88 & 81.55 \\
\midrule
\rowcolor{blue!20} \multirow{1}{*}{\textbf{Image + BBox}} 
& DLaVA \textbf{(Pixtral-12B)} & \textbf{85.91} & \textbf{66.96} & \textbf{76.34} & \textbf{87.57} & \textbf{84.41} & \textbf{91.42} \\
\bottomrule
\end{tabular}
}
\label{tab:free-anls}
\end{table*}

\section{Related Work}
\label{sec:related}

Recent advancements in multimodal document processing have significantly enhanced the capabilities of models in text detection, recognition, and information extraction. In this section, we review the relevant literature, focusing on the methods most pertinent to our work.

\subsection{Text Detection}



Accurate text detection is a foundational step for structured data extraction from unstructured documents. Recent methods have focused on improving accuracy and efficiency across various text orientations, sizes, and backgrounds. DBNet \cite{liao2020real} introduced a real-time differentiable binarization method that improved boundary localization while maintaining computational efficiency. FAST \cite{chen2021fast} further improved detection speed and accuracy for irregular text shapes, while MixNet \cite{zeng2023mixnet} utilized receptive fields and feature fusion to tackle complex scenes, marking significant strides in robust text detection.


    






\subsection{Text Recognition}
\label{sec:TextRecognition}


 Text recognition is an integral part of OCR-dependent approaches for document VQA as well as visual information extraction. In text recognition, the evolution from sequence models to transformer-based architectures has yielded models resilient to diverse fonts, distortions, and complex layouts. Early models such as CRNN \cite{shi2016end}, SAR \cite{li2019show}, and MASTER \cite{lu2021master} established the groundwork for sequence and attention-based recognition. More recent Transformer-based models, such as ViTSTR \cite{atienza2021vision} and PARSeq \cite{bautista2022scene}, further enhanced accuracy by capturing long-range dependencies. Innovations like MaskOCR \cite{lyu2022maskocr}, TrOCR \cite{li2023trocr}, and DTrOCR \cite{fujitake2024dtrocr} have integrated masked pretraining with encoder-decoder frameworks, achieving SoTA recognition accuracy across challenging scenarios.

\subsection{Information Extraction}
\label{sec:InformationExtraction}

Recent advancements in MLLMs have utilized both OCR-free and OCR-dependent architectures. OCR-free models, such as Donut \cite{kim2022ocr}, UDOP \cite{tang2023unifying}, and OmniParser \cite{wan2024omniparser}, bypass traditional OCR steps, reducing pipeline complexity and mitigating error propagation. Advanced OCR-free MLLMs, including Phi4 \cite{abdin2024phi}, LLaVAR \cite{zhang2023llavar}, Pixtral-12B \cite{agrawal2024pixtral}, Llama 3.2 \cite{dubey2024llama}, InternVL2 \cite{chen2023internvl, chen2024far}, Qwen2.5-VL \cite{bai2025qwen2}, LLaVA-Next \cite{liu2023improved}, and LLaVA-OneVision \cite{li2024llava}, extend multimodal comprehension, offering efficient extraction of structured data without dependency on external OCR processes.

In contrast, OCR-dependent models integrate OCR data to enhance document layout and positional comprehension. ICL-D3IE \cite{he2023icl} and LATIN-Prompt \cite{wang2023layout} incorporate positional data, though this can lead to increased input sequence length and slower inference. Recent approaches such as Cream \cite{kim2023visually} and InstructDoc \cite{tanaka2024instructdoc} streamline these processes by employing additional encoders to integrate OCR information, improving inference efficiency without compromising comprehension.

Despite these improvements, spatial precision and explainability remain challenging for document VQA applications. Our work addresses these challenges by introducing an integrated MLLM approach that merges text recognition and spatial understanding within a unified model, bypassing the need for separate OCR components and advancing spatial localization in document analysis.

\subsection{Layout-Aware Document Understanding}

Incorporating layout-specific information has proven effective in enhancing spatial comprehension in document understanding. LayoutLLM \cite{luo2024layoutllm} employs a layout instruction tuning strategy to improve the model's ability to interpret document layouts. DocLayLLM \cite{liao2024doclayllm} encodes OCRed textual, visual, and positional information directly within the model, removing the need for additional document encoders and refining comprehension through a CoT annealing process. LayTextLLM \cite{lu2024bounding} introduces a Spatial Layout Projector to convert OCR-derived coordinates into bounding box tokens, allowing seamless integration of spatial layouts with textual data. While these models enhance layout awareness, they often require complex adaptations or additional components that may affect model generality and increase computational overhead.

In summary, recent developments in multimodal document processing and layout-aware models have significantly advanced Document VQA capabilities, yet challenges in spatial precision, interpretability, trustworthiness and computational efficiency remain. These research gaps motivated our work, leading us to develop an innovative approach that addresses the challenges. 

%% file: sec/3_method.tex
\section{DLaVA}
\label{sec:method}

This section describes our proposed DLaVA approach for zero-shot, OCR-free information extraction from documents, as illustrated in Figure~\ref{fig:anot}. By harnessing the power of MLLM, our method directly extracts and localizes information from document images without relying on iterative OCR processing, thereby achieving robust structural accuracy while balancing computational efficiency with precise spatial grounding. The DLaVA approach is comprised of the following steps:

\begin{enumerate}
    \item \textbf{Text Detection Module}: The original document image \( I \) is processed using a text detection model—DB-ResNet-50 \cite{liao2020real}, as shown in step 1 in Figure \ref{fig:anot} as its real-time differentiable binarization method delivers superior boundary localization with high computational efficiency—a critical balance for structured data extraction in document images that is not as effectively achieved by FAST’s emphasis on irregular text shapes or MixNet’s complexity in handling intricate scenes.
    This model outputs bounding boxes for each text segment in the image. The detected bounding boxes are represented as:
    \[
    BB = \{ BB_1, BB_2, \dots, BB_n \}
    \]
    where each \( BB_i \) is a bounding box coordinate \( [x_{i1}, y_{i1}, x_{i2}, y_{i2}] \), labeled as step 2 in Figure \ref{fig:anot}.
    Each bounding box \( BB_i \) is used to crop a segment of the image \( I \), isolating individual words or phrases. The cropped image for \( BB_i \) is denoted by:
    \[
    C_i = I[BB_i]
    \]
    
    \item \textbf{Constructed Image Creation}: Instead of performing OCR on each cropped image, the bounding box images are arranged to form a “constructed image,” illustrated in step 3 of Figure \ref{fig:anot}. Each bounding box \( BB_i \) is assigned a unique ID for easy reference. The constructed image, \( I_C \), is an assembly where each line contains a cropped image, followed by its corresponding bounding box ID:
    \[
    I_C = \{ (C_1, BB_1), (C_2, BB_2), \dots, (C_n, B_n) \}
    \]
    For example, if the document contains sentences like ``THE STATE OF TEXAS...", after text detection, we obtain cropped images of individual words such as "THE" \((C_1)\), ``STATE" \((C_2)\), ``OF" \((C_3)\), and ``TEXAS" \((C_4)\). In the constructed image \(I_C\), each line would display the words with their bounding box IDs in sequence (e.g., the first line shows ``THE \((BB_1)\)", the second line ``STATE \((BB_2)\)", etc.).

    \item \textbf{Information Extraction Model}: In parallel, the MLLM - Pixtral-12B model \cite{agrawal2024pixtral} receives the input image \( I \) and the query \( Q \) (step 4) to generate the answer text \( A \). The generated answers, together with their corresponding questions (\( Q \)+\( A \)), are passed as an input to the final MLLM.

    \item \textbf{Final MLLM Processing}: 
    In the final step (step 5), the Pixtral-12B model utilizes the bounding box coordinates from step 2, the constructed image \( I_C \) from step 3, and the question-answer pair from step 4 to generate the answer's bounding box \( B_A \) and return it along with the answer \( A \). Subsequently, post-processing scripts are applied to annotate the returned answer based on the coordinates of \( B_A \).

\end{enumerate}

\noindent \textbf{Handling Cascading Errors:} 
Our approach avoids cascading errors by eliminating the explicit text recognition (OCR) step entirely. In traditional OCR-based systems, any misrecognition of text in the initial OCR stage propagates through subsequent stages, leading to errors in answer extraction and localization. In contrast, our method leverages an MLLM to directly extract and localize information from document images. We first detect text regions and then create a ``constructed image" that consolidates these regions along with their unique bounding box identifiers and corresponding coordinates. This unified representation is processed in one go—first to generate an initial answer and later to refine spatial localization—thereby bypassing the need for iterative OCR and preventing errors from accumulating. Furthermore, any inaccuracies introduced during the text detection phase are mitigated by the final MLLM (step 5), which leverages the overall contextual information to correct inconsistencies \cite{liu2024textmonkey}. This streamlined pipeline not only enhances accuracy but also improves computational efficiency and robustness in answer localization.

%% file: sec/4_experiment.tex
\begin{table}[htbp]
\centering
\setlength{\tabcolsep}{1pt}
\caption{Selecting best model for Text Recognition based on ANLS (for Ablation 3)}

\resizebox{\columnwidth}{!}{%
\begin{tabular}{lcccccc} 
\toprule
{\textbf{Models}} & \textbf{DocVQA} & \textbf{EST-VQA} & \textbf{RICO} & \textbf{FUNSD} & \textbf{CORD} & \textbf{SROIE} \\
\midrule
PARSeq \cite{bautista2022scene}   & \textbf{68.22} & 58.89         & \textbf{65.91} & \textbf{76.23} & \underline{77.21} & \underline{84.90} \\
MaskOCR \cite{lyu2022maskocr}     & 66.83       & 55.18         & 59.99        & 75.42       & \textbf{77.65}  & 83.38         \\
TrOCR \cite{li2023trocr}          & 64.86       & \underline{59.11} & 63.43        & 75.01       & 76.59         & 81.92         \\
DTrOCR \cite{fujitake2024dtrocr}  & \underline{67.93} & \textbf{60.08}  & \underline{63.77} & \underline{76.11} & 77.19         & \textbf{85.33} \\
\bottomrule
\end{tabular}
}
\label{tab:ablation3}
\end{table}

\section{Experiments}
\label{sec:Experiments}


\subsection{Datasets and Experimental Setup}

We evaluated our proposed model on several well-established, text-rich document datasets commonly used for VIE and Document VQA tasks. For VIE-related question answering, we utilized the \textbf{FUNSD}~\cite{jaume2019funsd}, \textbf{CORD}~\cite{park2019cord}, and \textbf{SROIE}~\cite{huang2019icdar2019} datasets. In the domain of Document VQA, we assessed performance using the \textbf{DocVQA}~\cite{mathew2021docvqa}, \textbf{RICO}~\cite{deka2017rico} datasets, and Scene Text+Evidence Visual Question Answering \textbf{(EST-VQA)}~\cite{wang2020general}. All models, including our proposed approach and baseline comparisons, were trained and evaluated on a single NVIDIA A100 GPU with 80\,GB of memory. This consistent computational environment ensures fair and reliable comparisons across different experimental settings.

We evaluated our model using two metrics to assess textual accuracy and spatial alignment, following established protocols. For textual accuracy, we used ANLS \cite{yujian2007normalized}, which measures normalized Levenshtein distance between predicted and ground truth answers, with values from 0 to 1 (1 indicating a perfect match). For spatial alignment, we employed IoU \cite{rezatofighi2019generalized}, which assesses overlap between predicted and ground truth bounding boxes. Performance was evaluated using mAP@IoU[0.50:0.95], where mean Average Precision (mAP) is computed across IoU thresholds from 0.50 to 0.95 in increments of 0.05. This metric captures the model’s ability to localize answer regions accurately across varying levels of spatial precision, providing a comprehensive measure of answer correctness and localization.

\subsection{Hyperparameter Details}
\label{hyperparameters}

We set the hyperparameters for each component in our framework to achieve an optimal balance between model efficiency and accuracy. After rigorous experiments with various hyperparameter ranges, we determined the following combinations to be optimal for our model. The configurations for each module are as follows:

\begin{itemize}
  \item \textbf{Pixtral-12B Model Hyperparameters:}
    \begin{itemize}[label=-]
      \item We set \texttt{max\_tokens} to 128k to avoid truncation for large multi-modal prompts; this parameter can be adjusted within the range of 8k to 128k.
      \item The \texttt{temperature} is fixed at 0.1, which lies within the permissible range of 0.0 to 1.0.
      \item We use a \texttt{top-p} value of 1.0 to enforce greedy selection under these constraints, with the value allowed to vary between 0.0 and 1.0.
    \end{itemize}

\vspace{2mm}

  \item \textbf{DB Resnet-50:}
    \begin{itemize}[label=-]
      \item The binarization threshold (\texttt{bin\_thresh}) is set to 0.3, which is within the acceptable range of 0.1 to 0.9.
      \item The box threshold is fixed at 0.1, and it may vary between 0.1 and 0.9.
    \end{itemize}

\vspace{2mm}

  \item \textbf{PARSeq:}
    \begin{itemize}[label=-]
      \item The maximum sequence length for positional embeddings (\texttt{max\_length}) is set to 32, and it can be adjusted between 16 and 256.
      \item All other hyperparameters for PARSeq remain at their default values.
    \end{itemize}

\vspace{2mm}
    
  \item \textbf{Hyperparameter Optimization:} We employed Optuna for hyperparameter optimization, and the final values were selected based on the best performance on the validation set to ensure a robust balance between computational efficiency and model accuracy.
\end{itemize}

\noindent Additionally, details regarding our model prompting strategy and the associated input-output formatting are described in Appendix \ref{appendix:model_prompting}.

\subsection{Baseline Models}


To evaluate the effectiveness of our proposed approach, we compare it against a comprehensive set of baseline models that span both OCR-free and OCR-dependent paradigms. Our comparisons include SoTA OCR-free models such as Phi4-14B \cite{abdin2024phi}, PixTral-12B \cite{agrawal2024pixtral}, InternVL~v2-8B \cite{chen2023internvl, chen2024far}, Qwen2.5-VL~7B \cite{bai2025qwen2}, LLaVA-OneVision~(OV)~7B \cite{li2024llava}, LLaVA-NeXT-13B (Vicuna) \cite{liu2023improved}, and LLaMA~3.2-11B \cite{dubey2024llama}. These models represent the current frontier in directly processing document images without the need for explicit OCR pipelines. In addition, we also benchmark our method against leading OCR-dependent models, including LLaMA~2-7B-Chat \cite{touvron2023llama}, LLaMA~3-8B-Instruct \cite{dubey2024llama}, LayoutLLM-7B \cite{luo2024layoutllm}, DocLayLLM \cite{liao2024doclayllm}, and LayTextLLM \cite{lu2024bounding}. This diverse collection of baselines not only enables us to evaluate our model's performance in terms of textual accuracy and spatial localization but also highlights the benefits of our unified, zero-shot OCR-free approach over traditional OCR-dependent methods.

\subsection{Ablation Study}
\label{sec:abblations}

We conduct the following ablation experiments to assess the contributions of different components in our OCR-Free pipeline (Ablation 1 and 2) and also compare it with an OCR-dependent approach (Ablation 3):

    
    

\begin{itemize}
    \item \textbf{Ablation 1 - Additional Image Input:} In this experiment, we provide the original input image \( I \) as an extra input to the final MLLM model (step 5 in Figure~\ref{fig:anot}) along with the other input components. This helps us evaluate the impact of the full visual context on the model's performance in extracting and localizing answers.
    
    \item \textbf{Ablation 2 - Removal of Information Extraction:} Here, we remove the information extraction step (step 4) entirely, relying solely on the final MLLM (step 5) for both question-answering and generating the corresponding bounding boxes. This experiment isolates the contribution of the dedicated information extraction module and demonstrates its role in refining spatial localization and answer accuracy.
    
    
    

    \item \textbf{Ablation 3 - OCR-Dependent Approach:} For comparison, we consider an OCR-dependent model that incorporates a text recognition module (PARSeq \cite{bautista2022scene}) to convert cropped images into text. Table~\ref{tab:ablation3} compares the text recognition accuracy of several cutting-edge OCR models (PARSeq, MaskOCR, TrOCR, and DTrOCR) across multiple benchmark datasets, and we observe that PARSeq achieves overall higher accuracy, making it the preferred module for our experiments. In this approach, a text detection model (DB-ResNet-50) is first used to obtain the detected cropped images (step 3) along with their corresponding bounding box coordinates (step 2). The cropped images are then passed to the text recognition module (step 3*) to generate textual representations, and the outputs from the text recognition step—together with the bounding box information and the input question—are fed into the final MLLM (step 5) to generate the answer \( A \) and its bounding box \( BB_A \). Figure~\ref{fig:ocr-diag} shows the architecture of the OCR-dependent model. This ablation serves as a baseline to highlight the benefits of our unified OCR-free approach over traditional methods that rely on separate text recognition.

\end{itemize}

%% file: sec/5_result.tex
\section{Results and Discussion}
\label{sec:result}


\begin{figure*}[h]
    \centering
    \includegraphics[width=1\textwidth]{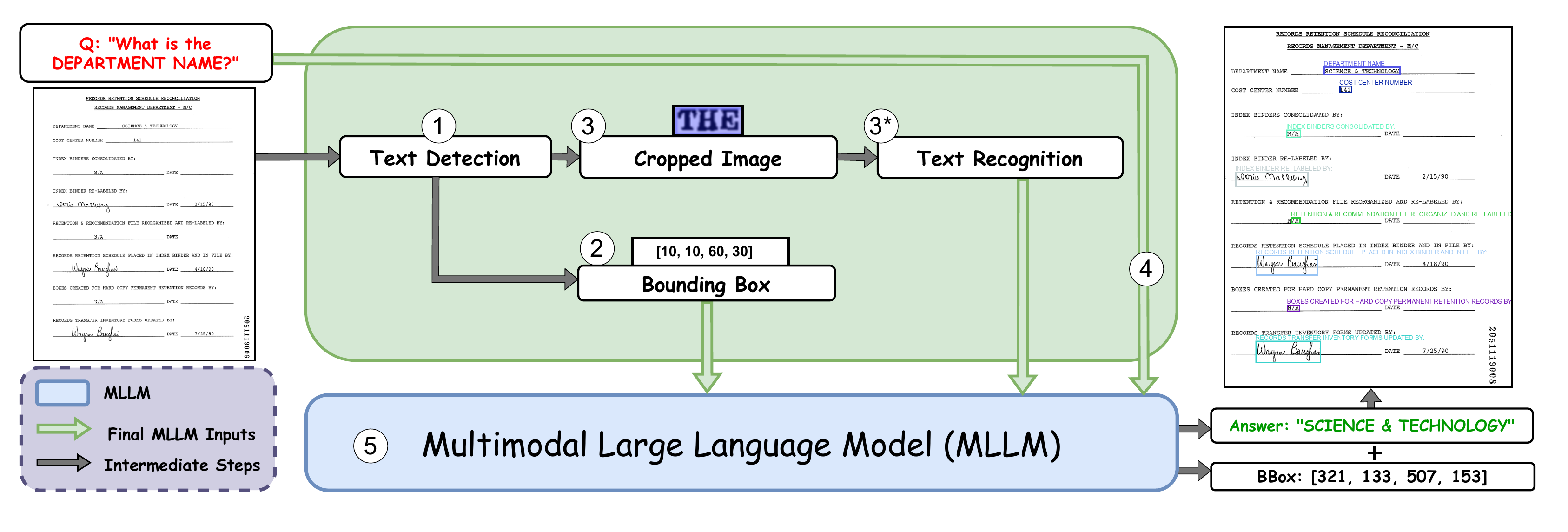}

    \caption{Architecture of OCR-Dependent Model (Ablation 3)}

    \label{fig:ocr-diag}
\end{figure*}

In this section, we present a comprehensive analysis of our proposed model’s performance compared to cutting edge baseline methods on both Document VQA and VIE tasks. We evaluate the model using standard metrics such as ANLS and IoU to capture both textual accuracy and spatial precision. In addition, we compare our model against several ablation variants to highlight the significance of each component in the overall pipeline.

\begin{table}[htbp]
\centering
\setlength{\tabcolsep}{3pt}
\caption{Ablation Study Results: Comparison of DLaVA and its ablation variants using ANLS metric on Doc. VQA \& QA for VIE}

\resizebox{\columnwidth}{!}{%
\begin{tabular}{lcccccc}
\toprule
\textbf{Models} & \textbf{DocVQA} & \textbf{EST-VQA} & \textbf{RICO} & \textbf{FUNSD} & \textbf{CORD} & \textbf{SROIE}  \\

\midrule
\rowcolor{blue!20} 
DLaVA   & \textbf{85.91} & \textbf{66.96} & \textbf{76.34} & \textbf{87.57} & \textbf{84.41} & \textbf{91.42}  \\

Ablation 1  & 83.55 & 64.01 & 69.41 & 83.28 & 79.08 & 85.36  \\
Ablation 2  & 82.26 & 62.51 & 73.86 & 84.35 & 81.91 & 86.02  \\
Ablation 3  & 74.02 & 62.70 & 71.99 & 79.57 & 82.08 & 90.45 \\
\bottomrule
\end{tabular}
}
\label{tab:ablation-free-anls}
\end{table}




\begin{table}[htbp]
\centering
\setlength{\tabcolsep}{5pt}
\caption{Ablation Study Results: Comparison of DLaVA and its ablation variants using IoU (mAP@IOU[0.50:0.95]) metric on Document VQA and QA for VIE}
\resizebox{\columnwidth}{!}{%
\begin{tabular}{lccccc}
\toprule
\textbf{Models} & \textbf{DocVQA} & \textbf{EST-VQA} & \textbf{RICO} & \textbf{FUNSD} & \textbf{CORD}  \\

\midrule

\rowcolor{blue!20} 
DLaVA  & \textbf{46.22} & \textbf{33.65} & \textbf{38.13} & \textbf{45.52}  & \textbf{57.86}   \\
Ablation 1  & 44.01 & 28.08 & 29.88 & 32.71 & 45.45  \\
Ablation 2  & 39.41 & 30.49 & 33.56 & 37.12 & 46.69 \\
Ablation 3  & 34.93 & 31.37 & 32.66 & 31.98 & 48.01 \\
\bottomrule
\end{tabular}
}
\label{tab:ablation-free-iou}
\end{table}



\subsection{Performance Analysis of DLaVA Models}
\label{OCR-Free}

We examine the performance of our proposed DLaVA model in comparison with existing baseline methods, using the ANLS metric on multiple benchmark Document VQA and VIE datasets, as summarized in Table~\ref{tab:free-anls}.

\begin{enumerate}
    \item \textbf{Document VQA Performance:} On the DocVQA dataset, DLaVA (Pixtral-12B) achieves an ANLS score of 85.91\%, outperforming previous SoTA approaches. Additionally, it obtains 66.96\% on the EST-VQA dataset and 76.74\% on the RICO dataset, demonstrating robust generalization across diverse document-question answering tasks. 
    EST-VQA is a bilingual dataset – it contains questions written in both English and Chinese, and for this dataset, as evident from Table~\ref{tab:free-anls}, other SoTA models struggle with multilingual recognition, resulting in suboptimal performance across various languages. In contrast, our DLaVA model delivers a consistent performance boost, providing at least around a 5\% improvement over the best SoTA models. This notable enhancement highlights the robustness and efficacy of our unified OCR-free approach in handling diverse linguistic inputs and reinforces its potential for superior multilingual document understanding.

\vspace{2mm}
    \item \textbf{VIE Task Performance:} DLaVA exhibits exceptional performance on various VIE tasks. On the FUNSD dataset, it achieves an ANLS score of 87.57\%, substantially surpassing baseline models. For the CORD dataset, it attains 84.41\%, and on the SROIE dataset, it scores 91.42\%, further underscoring its ability to handle different document layouts and textual complexities.
\end{enumerate}


\vspace{2mm}
In addition to these ANLS results, we also evaluate DLaVA’s spatial localization capabilities using the IoU metric (Table~\ref{tab:ablation-free-iou}) for the DocVQA, EST-VQA, RICO, FUNSD, and CORD datasets, where it obtains 46.22\%, 33.65\%, 38.13\%, 45.52\%, and 57.86\%, respectively. Although these IoU scores are lower than the corresponding ANLS scores, they offer valuable insights into the precision of bounding box alignment. Most existing approaches primarily focus on textual accuracy metrics (e.g., ANLS) while overlooking spatial correctness, which is critical for real-world applications requiring precise answer localization and interpretability. By reporting both ANLS and IoU, we gain a holistic view of the model’s performance, capturing both textual accuracy and spatial precision. Factors such as small fonts, stylized text, overlapping elements, and the inherent sensitivity of IoU to minor misalignments can lead to these discrepancies, underscoring the importance of considering spatial metrics alongside text-based evaluations.

DLaVA’s strong results can be attributed to multiple factors. First, by eliminating the need for a separate OCR module, it circumvents error propagation from text recognition, instead leveraging the visual-language capabilities of the MLLM through a constructed image that consolidates text regions with unique bounding box identifiers. Second, incorporating bounding box coordinates directly into the pipeline enhances spatial reasoning, enabling more precise answer localization. Although there remains room for improvement in fine-grained alignment, the bounding box integration significantly contributes to the model’s document layout understanding. Moreover, consolidating all identified text regions into a single constructed image reduces the overall context length, thus optimizing the MLLM’s processing efficiency. Operating in a zero-shot learning paradigm, DLaVA readily adapts to diverse document types and structures without additional training, ultimately excelling in both textual recognition and spatial localization for document understanding.

\subsection{Ablation Study Results}

We evaluate our proposed DLaVA model alongside three ablation variants (Ablation 1, Ablation 2, and Ablation 3 as described in Section~\ref{sec:abblations}) on Document VQA and VIE tasks using the ANLS metric (see Table~\ref{tab:ablation-free-anls}). DLaVA achieves the highest ANLS scores across all evaluated datasets, demonstrating its ability to accurately extract and interpret text information without reliance on iterative OCR. In contrast, Ablation 1 and Ablation 2—where either the original image was added as input or the information extraction step is removed—show reduced ANLS performance, underscoring the importance of both components for boosting textual accuracy and overall model effectiveness. Ablation 3, which incorporates an OCR-dependent process, also exhibits lower ANLS scores, indicating that our zero-shot OCR-free design is more robust to potential errors introduced by text recognition.

\vspace{2mm}

Table~\ref{tab:ablation-free-iou} reports the IoU scores for the same set of ablation experiments, focusing on bounding box localization. DLaVA again outperforms all ablation variants, reflecting its stronger spatial grounding capabilities. In particular, removing the dedicated information extraction step or excluding the original image input leads to noticeably lower IoU scores, highlighting how these design choices facilitate more precise bounding box predictions. Meanwhile, Ablation 3’s reliance on an external OCR stage can introduce cascading localization errors, resulting in lower IoU. 

\vspace{2mm}
Taken together, the ANLS and IoU metrics offer a holistic perspective: DLaVA not only excels in text accuracy but also in spatial alignment, affirming the benefits of its two-stage pipeline and constructed-image approach.

\subsection{Trustworthiness, Interpretability, and Explainability of the Proposed DLaVA Model}

Trustworthiness is a cornerstone of the DLaVA model’s design, fostering user confidence through its ability to deliver reliable and verifiable outputs in Document VQA. The model ensures precise answer localization, allowing users to directly verify responses by inspecting the corresponding document areas. This spatial grounding is further reinforced by robust evaluation metrics such as ANLS for textual accuracy and IoU for localization precision, which together establish the model as a dependable tool, reducing uncertainty and enhancing its practical utility. 
Moreover, as reliance on large MLLMs grows, so do concerns about AI hallucinations—instances where models generate plausible yet incorrect responses that can undermine user trust.
DLaVA addresses this critical challenge by coupling answer generation with explicit spatial annotations, thereby pinning each answer to a specific region in the document. This direct linkage not only enables users to verify the correctness of the output, but also significantly reduces the risk of hallucination, ultimately bolstering the overall trustworthiness and reliability of the system.

Building on this foundation, interpretability in the DLaVA model emerges from its transparent and modular architecture, which makes its operational process accessible. The pipeline’s distinct stages—such as text detection with DB-ResNet-50 followed by the constructed image—are clearly delineated, enabling step-by-step analysis of how inputs are transformed into outputs. This modularity, paired with the visual organization of text regions, provides a straightforward way to understand the model’s internal mechanics without relying on the specifics of its outputs or metrics.

Finally, explainability extends the model’s transparency by offering insights into its decision-making process, illuminating how conclusions are reached. During response generation, the model references specific bounding boxes, creating a traceable link between answers and their sources within the document. This capability, combined with the interplay of complementary performance indicators, sheds light on the model’s reasoning, though the inherent complexities of MLLMs may still pose challenges to achieving full clarity in every instance.

\section{Conclusion and Future Work}
\label{sec:conclusion}

In this paper, we presented DLaVA, a novel document language model that redefines visual question answering in documents through a unified, zero-shot OCR-free framework. By leveraging a two-stage MLLM pipeline and a uniquely constructed image—comprising cropped text regions annotated with distinct bounding box identifiers and corresponding coordinates—DLaVA bypasses traditional OCR processes and the cascading errors they entail. This innovative design not only reduces token overhead and computational complexity but also enhances spatial grounding and interpretability, as demonstrated by our SoTA performance on benchmark datasets such as DocVQA. Our experimental results confirm that DLaVA achieves superior textual accuracy and robust localization, thereby setting a new standard for reliability and transparency in document understanding. We believe that the streamlined integration of direct answer extraction with precise spatial annotations paves the way for more trustworthy and efficient AI systems in high-stakes applications.


Looking ahead, we plan to tackle challenges associated with lower IoU scores by refining bounding box annotations using fine-tuning techniques such as LoRA\cite{hu2021lora}, LoRA+\cite{hayou2024lora+}, QLoRA\cite{dettmers2024qlora}, and DoRA\cite{liu2024dora}. Furthermore, we aim to extend our approach beyond text localization to include embedded images, tables, and other structured or semi-structured content. By broadening the scope of DLaVA, we anticipate significant advancements in document understanding and the development of a more comprehensive solution for multimodal information extraction.

%% file: sec/append.tex
\clearpage
\setcounter{page}{1}

\appendix

\section*{Appendix}
\label{sec:appen}







\section{Examples of Ground Truth Answer Annotations}
\label{appendix:Examples-Ground-Truth-Answer-Annotations}

Appendix \ref{appendix:Examples-Ground-Truth-Answer-Annotations} presents some examples of ground truth annotations from the CORD and FUNSD datasets. These examples illustrate how document understanding tasks handle diverse document formats and content types. 

\begin{figure*}[h!]
    \centering
    \begin{subfigure}{0.49\textwidth}
        \includegraphics[width=\linewidth]{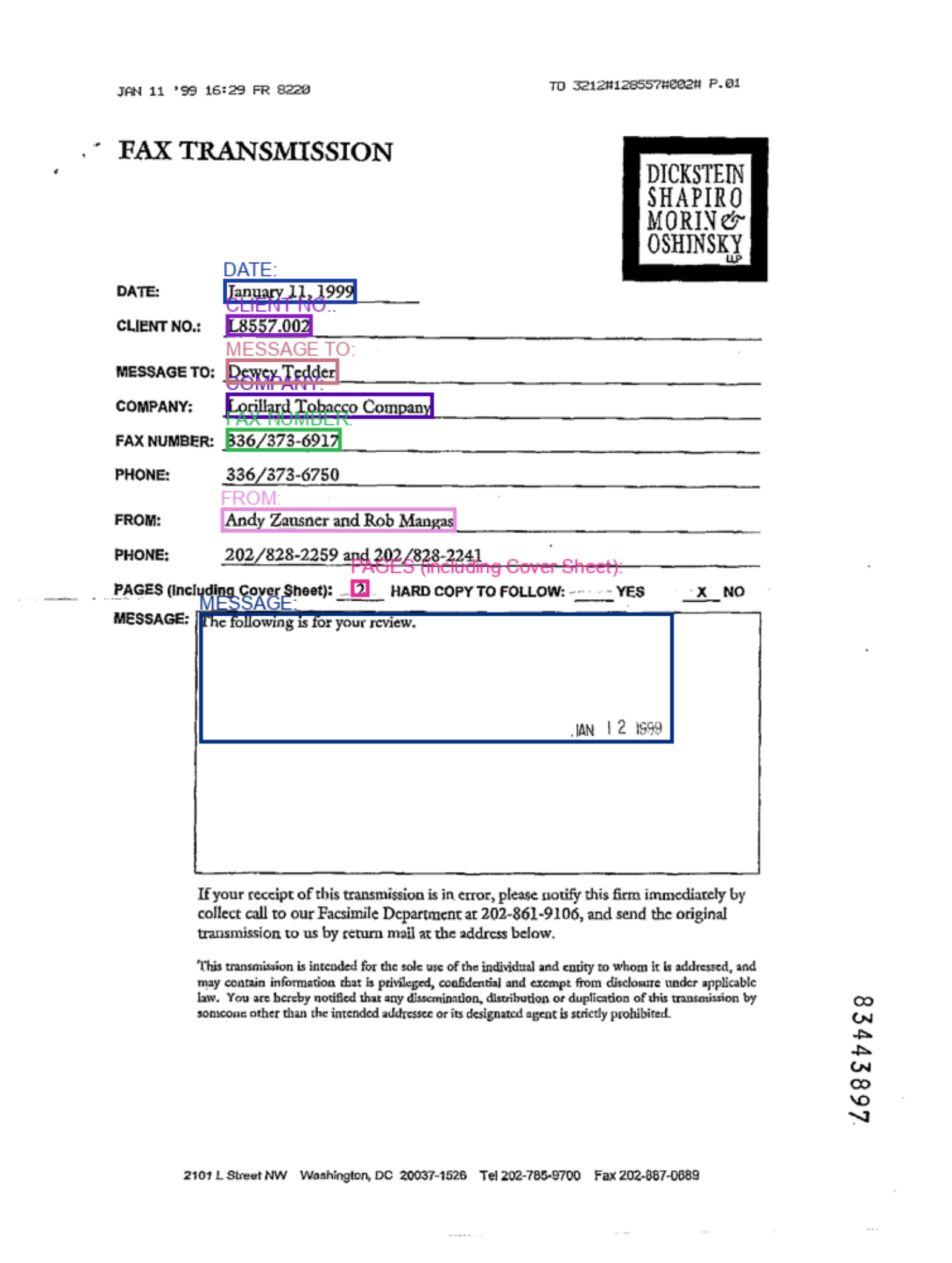}
        \caption{Document Example from FUNSD Dataset}
        \label{fig:FUNSD_HighANLS_lowIOU}
    \end{subfigure}
    \begin{subfigure}{0.40\textwidth}
        \includegraphics[width=\linewidth]{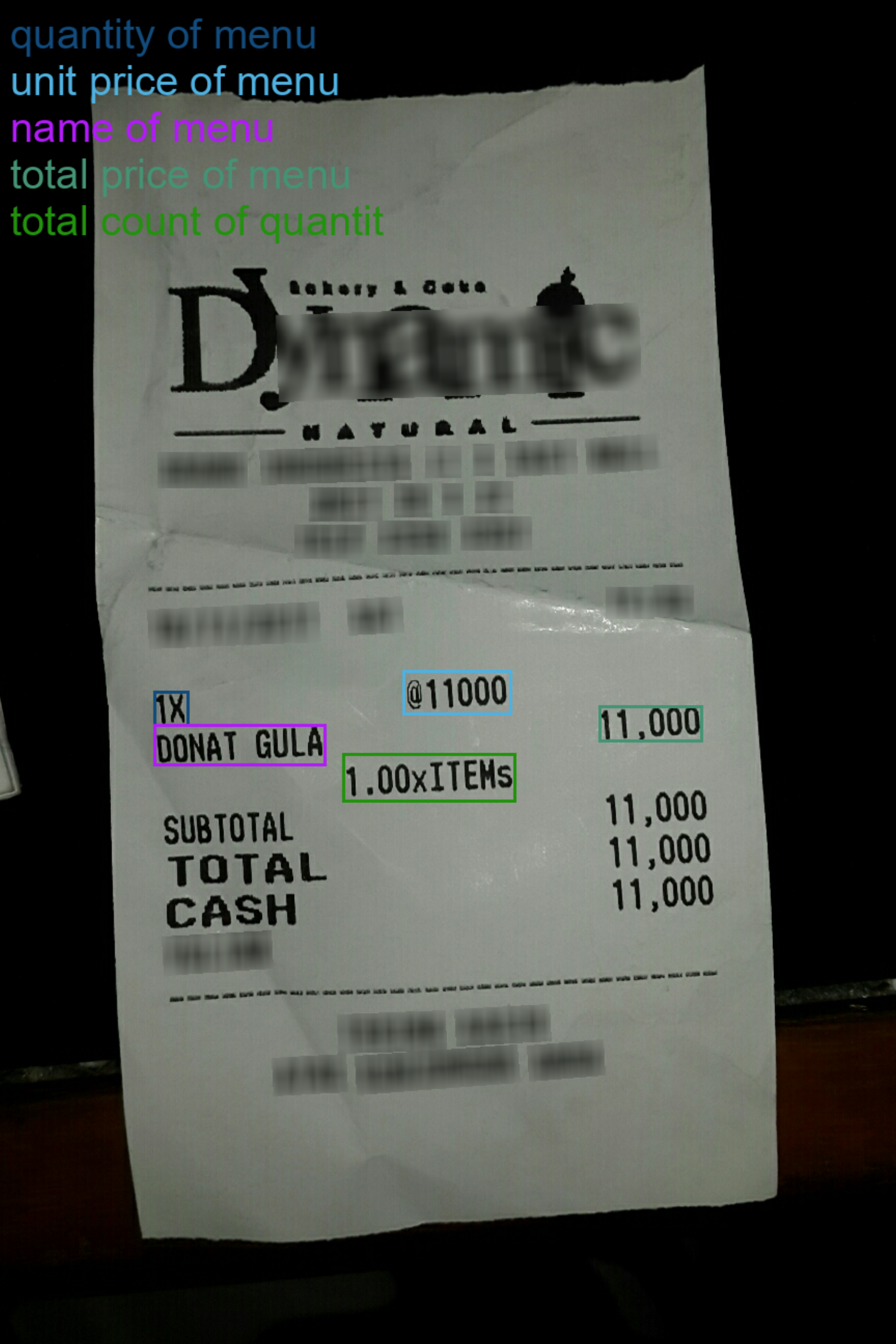}
        \caption{Receipt from CORD Dataset}
        \label{fig:CORD_HighANLS_lowIOU}
    \end{subfigure}
    
    \vspace{0.5cm}
    
    \begin{subfigure}{0.36\textwidth}
        \centering
        \includegraphics[width=\linewidth, trim={0 3.5cm 5cm 0}, clip]{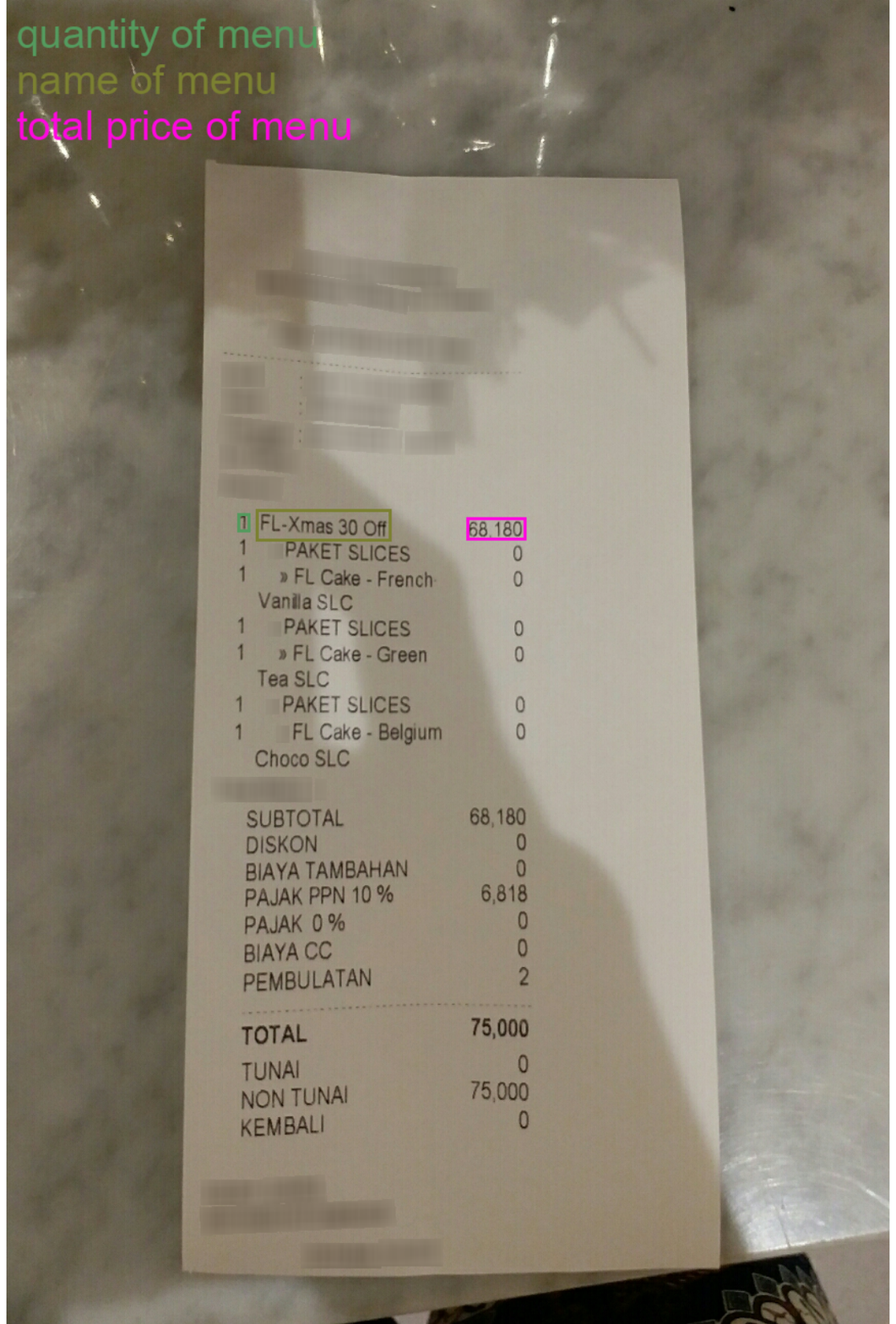}
        \caption{Another receipt from the CORD Dataset}
        \label{fig:CORD_HighANLS_highIOU}
    \end{subfigure}

    \caption{Illustrative Examples of Ground Truth Answer Annotations in Documents from the CORD and FUNSD Datasets}
\end{figure*}

Figure \ref{fig:FUNSD_HighANLS_lowIOU} depicts a document example from the FUNSD dataset, showcasing the structured layout of annotated key-value pairs in a form-like document. It highlights the ability to capture complex relationships between fields, such as dates, phone numbers, and textual descriptions.

Figure \ref{fig:CORD_HighANLS_lowIOU} displays a receipt example from the CORD dataset, emphasizing the annotation of essential receipt components like item quantity, unit price, total amount, and item names. This example underscores the importance of annotating critical transactional information typically found in unstructured receipt data.


Figure \ref{fig:CORD_HighANLS_highIOU} demonstrates another similar receipt from the CORD dataset. 


\section{Examples of Predicted Answer Annotations}
\label{appendix:Examples-Predicted-Answer-Annotations}
Appendix \ref{appendix:Examples-Predicted-Answer-Annotations} presents the answers and annotations generated by our proposed model, DLaVa (OCR-Free), for the same documents discussed in Appendix C. These examples provide insights into the model's ability to handle diverse document formats, such as structured forms and unstructured receipts, without relying on OCR. The illustrations highlight how DLaVa identifies key information and maps it to corresponding document regions, showcasing both its strengths and limitations. For example, the model demonstrates high semantic accuracy in extracting answers, as reflected in high ANLS scores, but sometimes struggles with precise spatial alignment, leading to lower IoU scores in some cases. By comparing these predictions with the ground truth annotations in Appendix \ref{appendix:Examples-Ground-Truth-Answer-Annotations}, readers can better understand the model's performance and areas for improvement.\\

\begin{figure*}[h!]
    \centering
    \begin{subfigure}{0.49\textwidth}
        \centering
        \includegraphics[width=\linewidth]{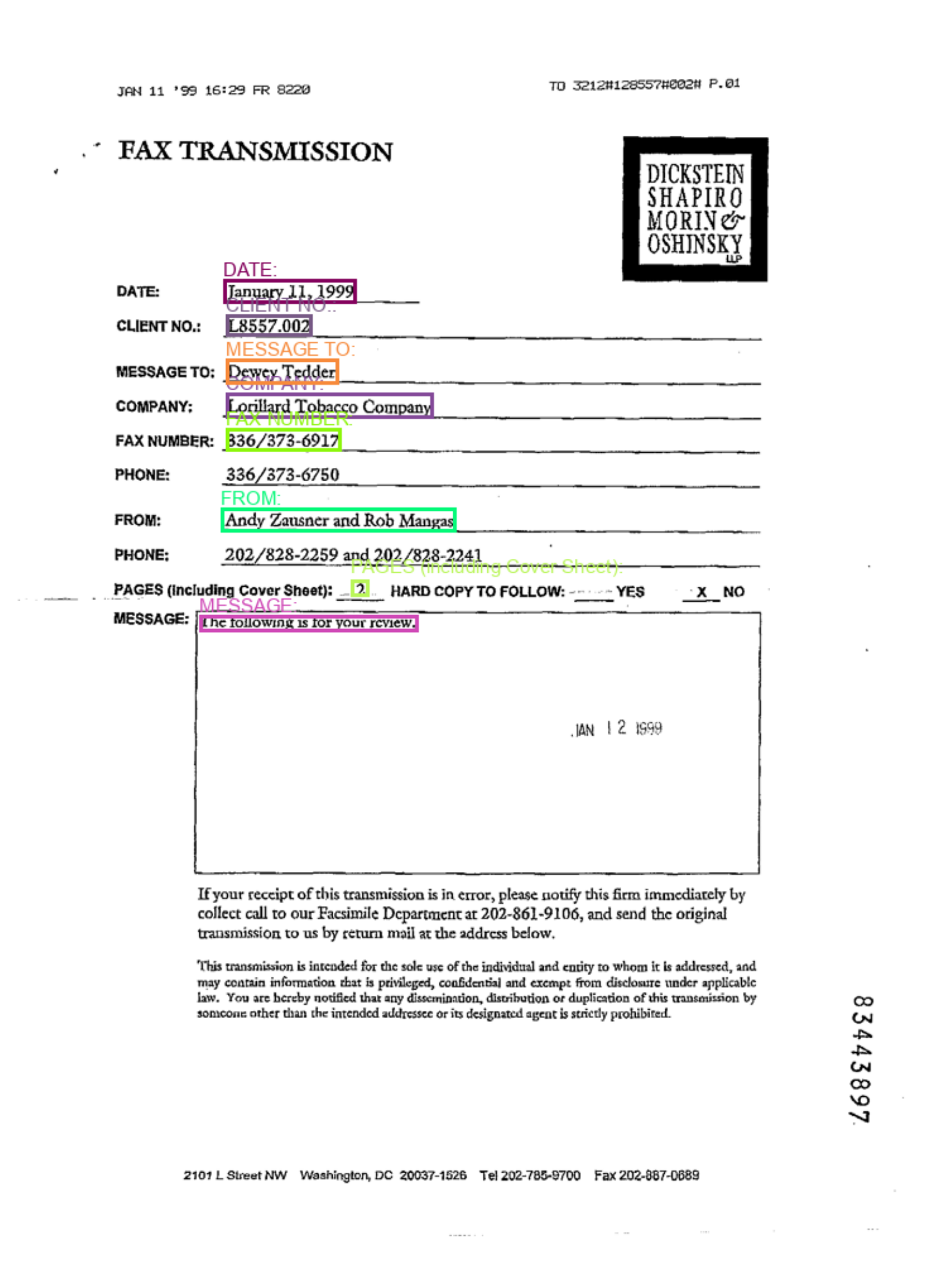}
        \caption{FUNSD-high ANLS, Low IOU}
        \label{fig:GT_FUNSD_highANLS_lowIOU}
    \end{subfigure}
    \begin{subfigure}{0.40\textwidth}
        \includegraphics[width=\linewidth]{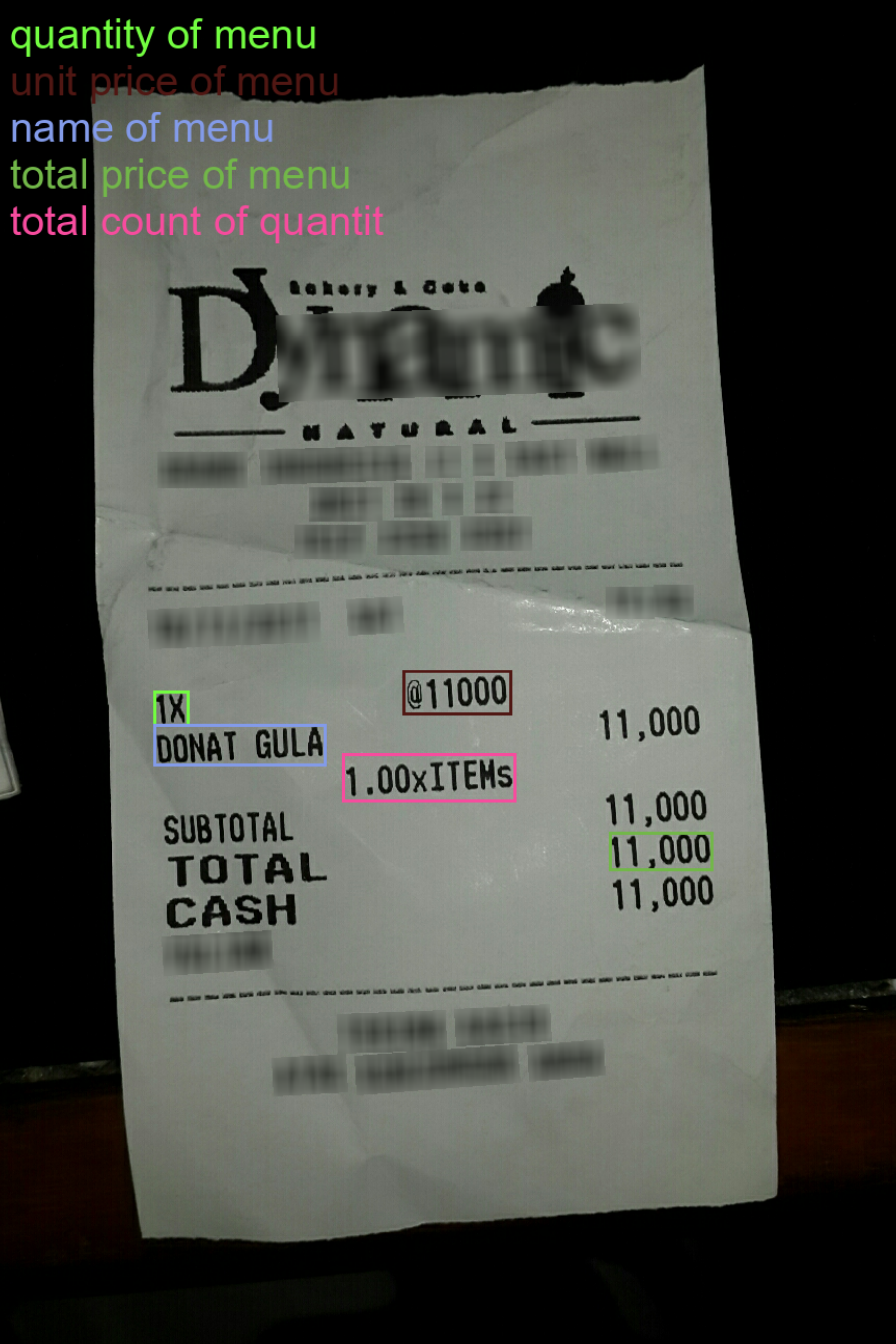}
        \caption{CORD-high ANLS, Low IOU}
        \label{fig:GT_CORD_highANLS_lowIOU}
    \end{subfigure}
    
    

    \vspace{0.2cm}
    \begin{subfigure}{0.36\textwidth}
        \centering
        \includegraphics[width=\linewidth, trim={0 5.5cm 5.5cm 0}, clip]{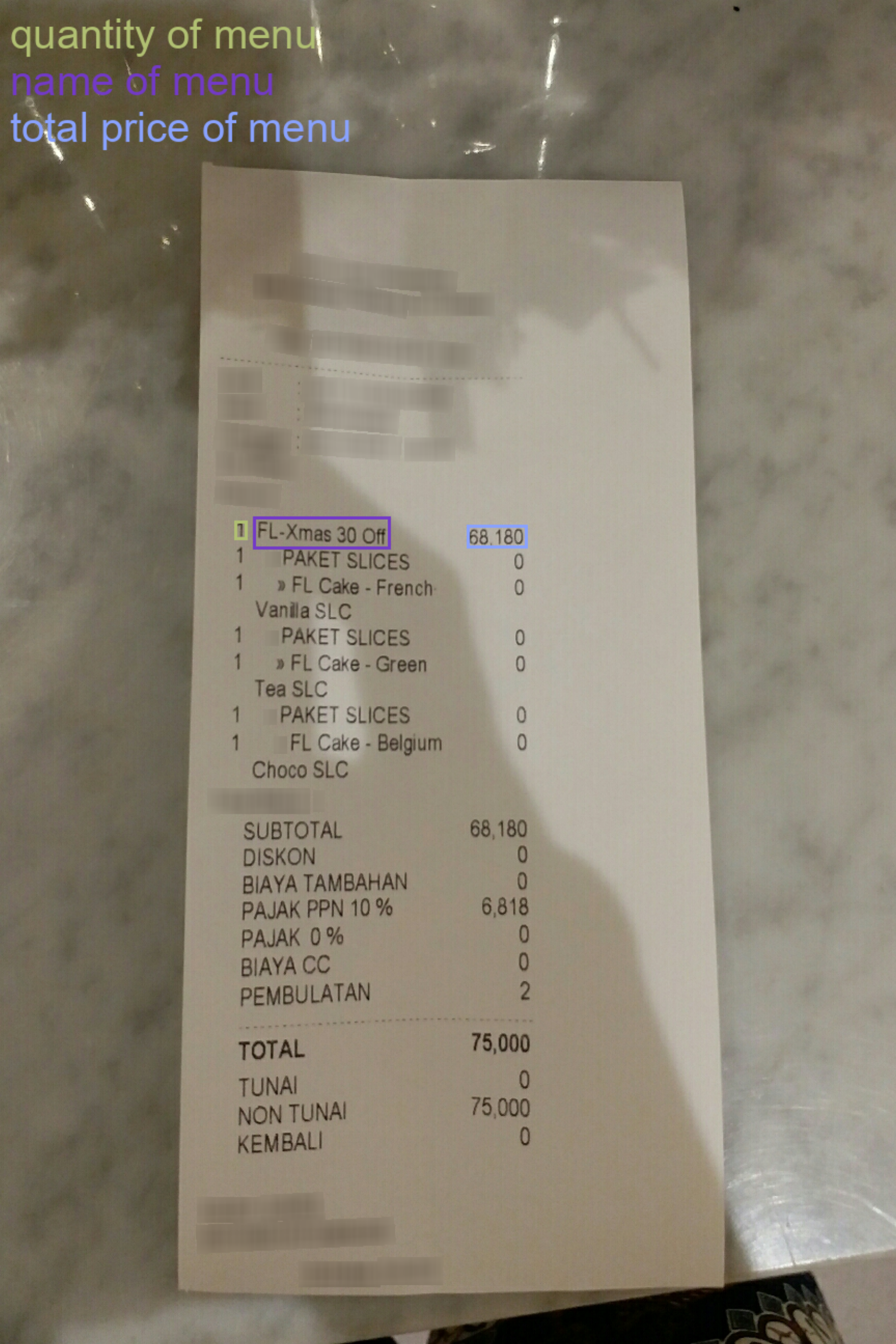}
        \caption{CORD-high ANLS, High IOU}
        \label{fig:PD_CORD_HighANLS_highIOU}
    \end{subfigure}

    \caption{Examples of Predicted Answer Annotations in Documents from the CORD and FUNSD Datasets}
\end{figure*}

Figure \ref{fig:CORD_HighANLS_highIOU}  shows a sample document where both the answers and their locations were identified with high precision by our model (as shown in Figure \ref{fig:PD_CORD_HighANLS_highIOU}). This resulted in an ANLS score of 100\% and an IoU nearly 100\%, as the model accurately captured the ground truth information.\\

Analysis for low IoU score between predicted and ground truth annotations for some cases:\\

\begin{enumerate}

    \item First, let us analyze a sample from FUNSD dataset. Figure \ref{fig:FUNSD_HighANLS_lowIOU} shows the ground truth answers for this sample along with their annotations, and Figure \ref{fig:GT_FUNSD_highANLS_lowIOU} shows the answers and annotations returned by our model DLaVa (OCR-Free) for the same document. 
    
    The IoU score for the ``Message" field of this document was observed to be 5.89\%, despite achieving a high ANLS score of 70.73\%. This discrepancy can be attributed to the differing interpretation of the message's spatial extent between the ground truth (Figure \ref{fig:FUNSD_HighANLS_lowIOU}) and the predicted annotations (Figure \ref{fig:GT_FUNSD_highANLS_lowIOU}).


    In the ground truth annotation, the bounding box includes the specific textual region containing the date component (``Jan 12, 1999") within the broader message context, towards the end of the box. However, our model's prediction restricts the bounding box to the ``Message" content, omitting the date. This misalignment results in a smaller predicted bounding box compared to the ground truth, thereby reducing the overlap and, consequently, the IoU score.


    This outcome highlights a common challenge in document understanding tasks, where predicted annotations may fail to encapsulate all semantically relevant content included in the ground truth. The low IoU score does not necessarily imply poor semantic accuracy but instead reflects a divergence in bounding box definitions.\\

    \item Let us analyze another sample from the CORD dataset. Figure \ref{fig:CORD_HighANLS_lowIOU} shows the ground truth answers for this sample along with their annotations, and Figure \ref{fig:GT_CORD_highANLS_lowIOU} shows the answers and annotations returned by our model DLaVa (OCR-Free) for the same document. 
    
    Here, in the task of extracting the ``Total Price of Menu" from receipt images, we observed that the IoU score was 0\%, despite achieving a perfect ANLS score of 100\%. This mismatch highlights an important limitation in the spatial alignment of predicted bounding boxes with the ground truth.


     In this instance, the value ``11,000" appears multiple times in the document, corresponding to different semantic fields (e.g., item price, subtotal, total price). While the model successfully identified the correct value for the ``Total Price of Menu," it incorrectly annotated a bounding box around the ``11,000" value associated with the total price of receipt rather than the ground truth location of the ``11,000" value corresponding to the total price of the menu. This resulted in no overlap between the predicted and ground truth bounding boxes, leading to an IoU score of 0\%.

    This case illustrates a common challenge in structured document understanding tasks where identical values appear in different semantic contexts. Resolving such issues requires incorporating additional contextual understanding into the model to ensure that annotations are correctly aligned with the intended semantic field. 
    As a part of the future work, we plan to explore incorporating positional priors, cross-field dependencies, or explicit disambiguation mechanisms to improve alignment between predictions and ground truth annotations.\\

\end{enumerate}


\section{Model Prompting Details} \label{appendix:model_prompting}

The model is prompted using the following inputs:
\begin{enumerate}
    \item A set of questions.
    \item The original input image where the answers to the questions are located.
    \item A JSON file containing the Bounding Box IDs (e.g., \texttt{BB0}, \texttt{BB1}, etc.) along with their corresponding bounding box coordinates for each word in the original input image.
    \item A second image displaying all words from the original input image along with their associated Bounding Box IDs.
\end{enumerate}

The prompt provided to the model is structured as follows:

\begin{lstlisting}[breaklines=true]
Questions:
{user_queries}

Bounding Box IDs and Bounding Box Coordinates for each word:
{bounding_boxes}

When finding answers to the questions, you are STRICTLY allowed to answer only using words present in the image. So, just return the words from the image (AND no description of full sentences). 
Just match the words that answer the question.

Your task is to find the answer to these questions from the 1st image, and identify the Bounding Box Coordinates for each answer. 
Return a JSON in the format specified below. (NO Additional Information. JUST JSON in the following format)

Final Answer: <answer>

where <answer> strictly adheres to the following structure:
- <answer> should be in JSON format.
- Each question from the question-answer pairs will be a key.
- For each question:
    - "value": The answer text (containing only words found in the input image; avoid point-wise or list-style answers).
    - "bounding_box": [[0.3037, 0.4863], [0.3257, 0.502]] (The bounding box coordinates in this exact structure).
      (Ensure only numerical digits, no NULL or empty values, and each coordinate is separated by commas).

If the answer consists of multiple words:
- Use the following format for "bounding_box":
    "value": "1 BLACK SAKURA" 
    "bounding_box": [
        [[0.09716796875, 0.458984375], [0.22314453125, 0.4921875]],
        [[0.23486328125, 0.462890625], [0.37548828125, 0.4873046875]],
        [[0.38720703125, 0.4619140625], [0.5556640625, 0.4873046875]]
    ]

Additional Instructions:
- Ensure correct pairing and matching of brackets (i.e., (), \{\}, []).
- Each "bounding_box" must contain exactly four numerical values formatted as two sets of coordinates within square brackets.

\end{lstlisting}